\documentclass[journal,twoside,hidelinks,a4paper]{IEEEtran}

\newcommand{\subparagraph}{}

\usepackage{emptypage,titling,titlesec,widetext,textcomp,relsize,parskip}
\usepackage[utf8]{inputenc}
\usepackage[margin=10pt,font=footnotesize,labelsep=period]{caption}
\usepackage{tabulary,longtable,makecell,multirow,color,colortbl}
\usepackage[colorlinks=false,linkcolor=black,urlcolor=gray,citecolor=black]{hyperref}
\usepackage[usenames,dvipsnames,svgnames,table]{xcolor}
\usepackage{amsmath,amssymb,amsfonts,amsthm}
\usepackage{authblk,lipsum,txfonts,float,morefloats,gensymb,graphicx,setspace,fancyhdr,booktabs,array,marvosym,xhfill}
\usepackage[normalem]{ulem}
\usepackage[export]{adjustbox}
\usepackage[includeheadfoot,paperwidth=8.5in,paperheight=11in,top=1cm,bottom=2cm,left=2cm,right=2cm,footskip=.25in,headheight=40pt,headsep=12pt,heightrounded]{geometry}

\newcommand{\ltf}{\mathrm{LT\mbox{-}F1}}
\newcommand{\ind}{\mathbf{1}}
\newcommand{\phead}[1]{\par\vspace{3pt}\noindent\textbf{#1}\hspace{0.5em}}
\theoremstyle{plain}
\newtheorem{theorem}{Proposition}
\newtheorem{lemma}{Lemma}

\renewcommand{\figurename}{Figure}

\titlespacing\section{0pt}{8pt plus 4pt minus 2pt}{2pt plus 2pt minus 2pt}
\titlespacing\subsection{0pt}{8pt plus 4pt minus 2pt}{2pt plus 2pt minus 2pt}
\titlespacing\subsubsection{0pt}{8pt plus 4pt minus 2pt}{2pt plus 2pt minus 2pt}

\newcommand{\subtitle}[1]{\posttitle{\par\end{center}\begin{center}\bfseries\Large #1\end{center}\vskip1.2em}}
\newcommand\myabstract[2][.8]{\renewcommand\maketitlehookd{\mbox{}\medskip\par\centering\begin{minipage}{#1\textwidth}\small#2\end{minipage}}}

\definecolor{myred}{RGB}{189, 3, 34}
\date{}

\title{Selective, Risk-Controlled Emotion Forecasting in Conversation: From a Persistence Barrier to Departure Detection with Conformal Guarantees}

\subtitle{}

\author[1]{\bfseries Bin Wen}
\author[2]{\bfseries Dai-Qiao Zhang}
\author[1]{\bfseries Tien-Ping Tan}

\affil[1]{School of Computer Sciences, Universiti Sains Malaysia, 11800 Gelugor, Penang, Malaysia}
\affil[2]{School of Mathematical Sciences, Universiti Sains Malaysia, 11800 Gelugor, Penang, Malaysia}
\affil[ ]{\vspace{0pt}}
\affil[ ]{Corresponding author: Tien-Ping Tan (e-mail: tienping@usm.my)}
\affil[ ]{\vspace{0pt}}

\begin{document}

\myabstract[.92]{\vspace{-1cm}\hrule\vspace{2mm}\textbf{Abstract:} Forecasting emotion in conversation is increasingly framed as predicting the next emotional state's value. We show this targets a hard quantity in a state-observed testbed and propose a learnable, deployable reformulation instead. On valence-arousal-dominance trajectories from a dyadic corpus, repeating a speaker's own previous state sets a persistence baseline that no frozen-encoder correction meaningfully improves upon and a state-space filter exceeds; the floor requires the historical state to be \emph{observed}, since an estimated anchor collapses persistence below a direct forecaster. What survives is the \emph{timing} of a departure from persistence, learnable on three of four corpora and near chance on the highest-base-rate corpus, a negative result we report. We recast the task as selective departure forecasting: a lead-time-weighted metric with a leakage protocol, and commitment policies whose best choice switches with the departure rate. A dialogue-level conformal risk control bounds the expected loss of committed predictions, meeting the target in all twelve configurations tested, though the guarantee is bought through abstention, with coverage and alarm recall falling sharply on the hardest corpus. One tested language model's confidence fails as a ranking signal where another succeeds, a model-dependent failure not yet attributable to family versus scale. Code and result logs are available from the corresponding author upon request.
\newline
\newline
\textbf{Keywords:} affective computing, conformal prediction, emotion forecasting, selective prediction \vspace{1mm} \hrule}

\maketitle

\section{Introduction}
\label{sec:introduction}

Emotion forecasting in conversation (EFC) asks a model to commit to a prediction about an emotional state that is not yet observable -- the condition that separates it from the mature emotion-\emph{recognition} literature \cite{dialoguernn,dialoguegcn,dagerc,efcsurvey}. The emerging consensus formulation inherits recognition's target: predict the \emph{value} of the forthcoming state, whether a categorical label \cite{altarawneh2023,shi2024,lu2023} or a point in valence-arousal-dominance (VAD) space. The appeal is practical: a customer-service agent that anticipates a caller's escalating frustration can intervene before the call is lost; a mental-health support chatbot that anticipates an affective downturn can adjust its response before distress is expressed; a tutoring system that anticipates disengagement can re-engage a student pre-emptively. Across all three use cases, the underlying assumption is the same -- that the \emph{value} of the next emotional state is the right quantity to predict, and that a sufficiently capable model will eventually predict it well.

This paper reports a systematic, audited attempt to build such value forecasters, and its central finding is that -- within the state-observed testbed we study -- the field may be aiming at the wrong quantity. On continuous VAD forecasting on IEMOCAP, where a historical emotional-state value is available as the reference, a trivial rule -- repeat what the same speaker was feeling last time they spoke -- is not beaten by any frozen-encoder model we train, while a classical state-space (Kalman) baseline over the same anchor performs somewhat better. This result is explicitly conditional on \emph{observing} that historical state: when the anchor must instead be \emph{estimated} from raw content, its dominance disappears (Section~\ref{sec:barrier}), so we do not claim persistence is a good deployable value forecaster, only that residual value beyond an observed state is hard to predict. What is beatable, and more actionable, is a narrower question -- not \emph{what} the next emotional state will be, but \emph{whether} it is about to change -- learnable on three of the four corpora studied, with EmoryNLP providing a chance-level negative result. Redirecting the target from value to occurrence yields a task that is both empirically solvable and deployable with formal statistical guarantees, which we develop and validate in this paper. We are explicit throughout about the boundary of the value-forecasting claim: it rests on frozen-encoder architectures with correction heads up to 3.7M parameters, an observed-state reference, and continuous-VAD evaluation on IEMOCAP, and does not rule out that full end-to-end fine-tuning or substantially different architectures could close the gap (Section~\ref{sec:discussion}).

\phead{The Persistence Barrier.} On turn-level continuous VAD trajectories from IEMOCAP \cite{iemocap}, the strongest zero-parameter baseline is not the conventional ``repeat the previous turn'' but \emph{same-speaker persistence}: repeat the target speaker's \emph{own} previous state, skipping over the interlocutor's intervening turn. This baseline reaches concordance correlation coefficient (CCC) $0.625/0.609/0.552$ at forecast horizons $t{+}1/t{+}2/t{+}3$ -- far above the turn-level variant ($0.446/0.549/0.426$). A classical, causal, few-parameter same-speaker Kalman filter does slightly better still ($0.662/0.647/0.603$), while an AR(1) linear recalibration of the same anchor does not ($0.560/0.539/0.471$). No \emph{unconstrained} direct value forecaster we train approaches either classical baseline: the best one attains CCC $0.140$ at $t{+}1$. The controlled version of this comparison -- an architecture that structurally anchors its output on the same gold same-speaker label persistence uses and only learns a gated correction on top -- lands, unsurprisingly, close to the persistence floor by construction ($0.629$ versus $0.625$ CCC at $t{+}1$), and a dedicated ablation shows that gated correction contributes nothing measurable beyond what the anchor already provides; this apples-to-apples result, not the raw gap, is what the paper's learnability claim rests on. The gap for unconstrained models survives a $60\times$ capacity sweep and four feature configurations up to multimodal RoBERTa-large plus WavLM audio; a psychologically structured Ornstein-Uhlenbeck (OU) dynamics model \cite{oravecz2011}, also unconstrained, trails persistence by $6.5$--$11\times$. We name this the \textbf{Persistence Barrier}, sharpening the ``forecasting floor'' phenomenon reported for coarser affect series \cite{cardioaffect}.

\phead{The Crack in the Barrier.} The same sweeps that flatten value prediction leave exactly one signal intact: \emph{whether} a departure from persistence is about to occur. Departure \emph{occurrence} is predictable above base rate in fifteen of fifteen fold-model cells (AUPRC up to $0.305$ versus a base rate of $0.237$); departure \emph{magnitude and direction} are not -- direction accuracy at horizons $t{+}2/t{+}3$ falls to $0.42$--$0.44$, below chance. The informative target is the break, not the level, and across the three corpora we test its predictability shows a consistent monotone association with the departure base rate, strongest exactly where departures are rare -- the regime in which anticipation matters most.

\begin{figure}[t]
\centering
\includegraphics[width=0.98\linewidth]{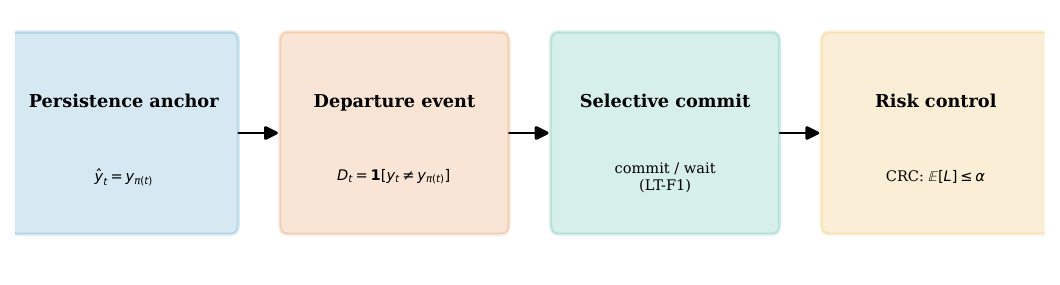}
\caption{Method overview: same-speaker persistence anchor $\to$ departure event $\to$ selective commitment $\to$ dialogue-level conformal risk control.}
\label{fig:schematic}
\end{figure}

\phead{From Forecasting to Selective Commitment.} Because departures are rare and hard to predict, a deployable forecaster must also decide \emph{when to speak} (Fig.~\ref{fig:schematic}). We study four commitment policies and find that none dominates: a cost-sensitive myopic policy wins at low departure base rates, a planning-based optimal-stopping policy wins at high rates, and a regime-adaptive meta-policy that estimates the local base rate matches the best single policy on every corpus tested. Reliability cannot be assumed: wrapping any policy in our dialogue-level conformal risk control (CRC) procedure yields a finite-sample guarantee that the \emph{expected} loss of \emph{committed} predictions on a test dialogue stays below a user-chosen risk level $\alpha$, which we verify empirically in all twelve primary configurations, while exposing the coverage cost at which the guarantee is bought.

\phead{Contributions.} (1) A unified \emph{selective departure forecasting} formalization across discrete and continuous affect representations, with a lead-time-weighted metric ($\ltf$) and a data-leakage test suite available from the corresponding author upon request. (2) The \emph{Persistence Barrier}: a triply audited negative result, within a frozen-encoder testbed, with a partial annotator-artifact control. (3) A study of commitment policies establishing base-rate-dependent optimality, and a regime-adaptive meta-policy that recovers the per-regime optimum without oracle knowledge of the base rate. (4) A dialogue-level \emph{instantiation} of conformal risk control for risk-controlled anticipation, together with an empirical analysis of the risk--coverage--recall trade-off it buys. (5) Three design principles distilled from audited negative results, including the double falsification of psychology-inspired dynamical priors as value forecasters and the model-dependent unreliability of LLM verbalized confidence as a ranking signal. All experiments run on a single consumer laptop; code, configuration files, and result logs are available from the corresponding author upon request.

\begin{figure*}[t]
\centering
\includegraphics[width=0.85\textwidth]{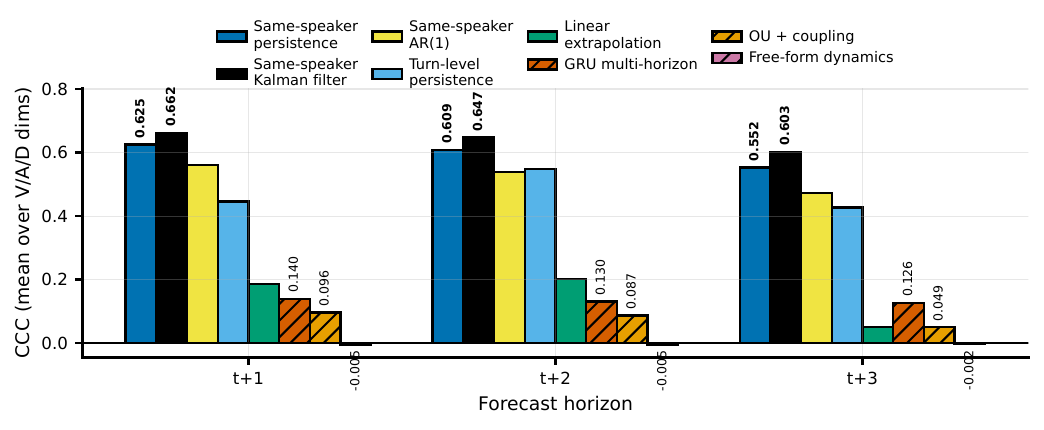}
\caption{The Persistence Barrier (IEMOCAP, 5-fold session-held-out, CCC averaged over valence/arousal/dominance). Solid bars: zero/few-parameter baselines; hatched bars: trained value forecasters. Both classical same-speaker baselines dominate every trained model at every horizon; free-form latent dynamics sits at essentially zero CCC.}
\label{fig:barrier}
\end{figure*}

\section{Related Work}
\label{sec:related}

A recent survey \cite{efcsurvey} maps EFC as a target-absent task distinct from recognition, shift-aware recognition, and flip reasoning, and calls for lead-time-aware evaluation and leakage reporting; we cite it for that framing, not as evidence for our own novelty. Existing forecasters predict next-turn labels or distributions from history \cite{bothe2017,wang2020nsf,li2020cell,zhang2021,altarawneh2023,shi2023,shi2024,lu2023,ju2023}, including implicit-reaction \cite{koga2024} and consequence \cite{ecfcon} variants; Hi-EF \cite{hief} provides a dedicated benchmark, and calibration has been studied for \emph{recognition} \cite{roohi2025} but not forecasting. None of this published prior work reports a lead-time-weighted metric, an executable leakage protocol, selective commitment, or a dialogue-level risk guarantee -- the pieces of evaluation and decision machinery this paper contributes on top of the value-vs-occurrence reformulation. Forecasting conversational derailment introduced unknown-horizon commitment \cite{craft} and, recently, an explicit ``when to trigger'' decision mechanism \cite{derailwait}; we transfer the commitment question to affective targets, replace heuristic rules with cost-sensitive, planning-based, and base-rate-adaptive policies, and add distribution-free risk control. Selective prediction has a long line of foundational work \cite{elyaniv2010,geifman2017}, with temperature scaling as the standard calibration baseline \cite{guo2017}. Conformal risk control (CRC) extends conformal prediction to bounding expected losses of thresholded predictors \cite{crc}, including a corollary for non-monotone losses via suffix-max monotonization; selective variants have recently been unified \cite{scrc}, and further non-monotone extensions are emerging \cite{nonmonocrc}. We do not claim new CRC theory: our contribution is its \emph{instantiation} on a \emph{dialogue-level} exchangeable unit with a ratio-type selective-forecasting loss and abstention convention, plus the empirical risk--coverage--recall analysis of what that guarantee costs. Psychology models emotion as an inertial, mean-reverting process with interpersonal coupling \cite{kuppens2010,oravecz2011,butler2011}; we test these priors computationally and report their double falsification for forecasting, while confirming the internal importance ordering of their components. Latent world models \cite{worldmodels} have recently been extended with affect \cite{affectverse,lewm,dreamcub}; none of this work addresses target-absent selective forecasting.

\section{Selective Departure Forecasting}
\label{sec:formal}

A conversation is a sequence of utterances $u_1,\dots,u_T$ with speakers $s_t$ and annotations $y_t$ (a categorical label, or a vector in VAD space). Let $C_t=(u_1,\dots,u_t)$. A \emph{forecast} at time $t$ is any prediction about $y_{t'}$, $t'>t$, computed from $C_t$ alone: no token, audio frame, or annotation from $u_{t'}$ may enter the input. All feature windows, normalization statistics, and thresholds are fit on training folds only; a test suite asserts this boundary programmatically (56 tests).

\phead{Is the future speaker's identity known?} The boundary above excludes future \emph{content} but is silent on future \emph{turn-taking structure}: our implementation assumes the identity $s_{t'}$ of the speaker at the target position is known at commit time $t$, since it selects which same-speaker anchor $\pi(t')$ the forecast is measured against. This is essentially free on DailyDialog (a strictly alternating two-party corpus), a real but modest assumption on the dyadic MELD subset used in our main tables ($25.6\%$ of turns there are followed by the same speaker), and most consequential on the full multi-party corpora: an order-2 Markov next-speaker predictor, fit on speaker sequences alone, reaches only $52.7\%$/$56.7\%$ top-1 accuracy on MELD/EmoryNLP test dialogues. We report this as an explicit scope boundary: this paper establishes that \emph{given} the future speaker's identity, predicting whether they will depart from their own persistence is learnable; forecasting jointly over an unknown future speaker and their departure is a harder problem we leave open.

Let $\pi(t)$ index speaker $s_t$'s most recent prior utterance. The \emph{same-speaker persistence} forecast is $\hat y_t=y_{\pi(t)}$. A \textbf{departure} is the event that reality deviates from it:
\begin{equation}
D_t=\ind[y_t\neq y_{\pi(t)}]\ \text{(discrete)},\quad
D_t^{(q)}=\ind[\lVert y_t-y_{\pi(t)}\rVert>\tau_q],
\label{eq:departure}
\end{equation}
with $\tau_q$ the $q$-th percentile of training-fold deviations (continuous case). This unifies discrete emotion shift and continuous change, and makes persistence -- not an arbitrary model -- the reference point a forecaster must beat. A selective forecaster is a pair $(f,\rho)$: a base model emitting $\hat P(D_{t'}\mid C_t)$ at each decision point, and a commitment policy $\rho$ that either \emph{commits} or \emph{waits}. Committing at $t$ against target $t'$ yields lead time $\Delta t=t'-t$.

\phead{Lead-Time-Weighted Evaluation.} Let TP be committed, correct departure predictions (each with lead $\Delta t_i$), FP committed false alarms, FN missed true departures, and $|C|=|\mathrm{TP}|+|\mathrm{FN}|$. With $w(\Delta t)=\min(\Delta t/\tau,1)$ and horizon $\tau=3$,
\begin{equation}
P=\frac{|\mathrm{TP}|}{|\mathrm{TP}|+|\mathrm{FP}|},\qquad
R_{LT}=\frac{1}{|C|}\sum_{i\in\mathrm{TP}}w(\Delta t_i),\qquad
\ltf=\frac{2\,P\,R_{LT}}{P+R_{LT}}.
\label{eq:ltf}
\end{equation}
A perfect but latest-possible forecaster ($\Delta t{=}1$) earns $w{=}1/3$; earlier correct commitments earn proportionally more, capped at $\tau$; indiscriminate alarming is penalized through $P$. On MELD, the optimal-stopping policy's advantage over the calibrated threshold is stable across $\tau\in\{2,3,5\}$ ($+3.6/+6.8/+6.7$ points). On DailyDialog it is not: the calibrated threshold edges out the meta-policy by $0.3$ points at $\tau{=}2$ (statistically indistinguishable), but the meta-policy leads by $1.7$--$1.8$ points at $\tau{=}3,5$.

\section{The Persistence Barrier}
\label{sec:barrier}

\phead{An input asymmetry.} The same-speaker persistence baseline's prediction \emph{is} the gold label $y_{\pi(t)}$. The unconstrained value forecasters it is compared against (GRU multi-horizon, OU-coupled dynamics, free-form latent dynamics) receive no such label history as input -- only frozen text/audio embeddings and a speaker identity per turn, never the numeric VAD trajectory itself. Part of the headline gap is therefore attributable to persistence having direct access to the exact quantity being predicted, one step removed. Audit~1 below is the controlled version -- a persistence-\emph{anchored} architecture given the gold anchor $y_{\pi(t)}$ directly, with only its \emph{additional} learned correction tested -- and it is this audit, not the raw gap, that the paper's learnability claim rests on.

\phead{What if the anchor itself has to be estimated?} Every number above assumes the anchor $y_{\pi(t)}$ is available as ground truth at commit time. We quantify this oracle assumption directly: a minimal same-turn probe (a small frozen-encoder regression head, no dialogue history) is trained to predict a turn's \emph{own} gold VAD from its \emph{own} content, and its output substituted for the gold anchor. The probe itself is weak (mean validation CCC $0.130$), and substituting it collapses same-speaker persistence's CCC from $0.625/0.609/0.552$ (oracle anchor) to just $0.071/0.069/0.058$ (estimated anchor, Table~\ref{tab:anchor-estimated}) -- \emph{below} every unconstrained trained model in Table~\ref{tab:barrier} except free-form dynamics. The entire Persistence Barrier result is therefore conditional on oracle access to the historical gold label, and Table~\ref{tab:barrier}'s framing should be read as \emph{state-observed} emotion forecasting, not as a claim about a fully realistic pipeline that must also recover that state from raw signal.

\begin{table}[t]
\caption{Same-speaker persistence, oracle vs.\ estimated anchor (IEMOCAP, 5-fold, mean CCC over V/A/D). ``Estimated'' substitutes a same-turn frozen-encoder probe's own output for the gold anchor value.}
\label{tab:anchor-estimated}
\centering
\begin{tabular}{@{}lccc@{}}
\toprule
Anchor & $t{+}1$ & $t{+}2$ & $t{+}3$ \\
\midrule
Oracle (gold label) & 0.625 & 0.609 & 0.552 \\
Estimated (same-turn probe) & 0.071 & 0.069 & 0.058 \\
\bottomrule
\end{tabular}
\end{table}

\begin{table}[t]
\caption{The Persistence Barrier (IEMOCAP, 5-fold session-held-out, CCC over valence/arousal/dominance).}
\label{tab:barrier}
\centering
\begin{tabular}{@{}lccc@{}}
\toprule
Predictor & $t{+}1$ & $t{+}2$ & $t{+}3$ \\
\midrule
\emph{Same-speaker persistence (0 par.)} & 0.625 & 0.609 & 0.552 \\
Same-speaker Kalman filter (1 hyperpar.) & \textbf{0.662} & \textbf{0.647} & \textbf{0.603} \\
Same-speaker AR(1) (2 par.) & 0.560 & 0.539 & 0.471 \\
Turn-level persistence (0 par.) & 0.446 & 0.549 & 0.426 \\
Linear extrapolation (0 par.) & 0.186 & 0.201 & 0.051 \\
\midrule
\multicolumn{4}{@{}l}{\emph{Unconstrained direct value forecasters:}} \\
GRU multi-horizon & 0.140 & 0.130 & 0.126 \\
OU + interpersonal coupling & 0.096 & 0.087 & 0.049 \\
Free-form latent dynamics & $-0.005$ & $-0.005$ & $-0.002$ \\
\midrule
\multicolumn{4}{@{}l}{\emph{Persistence-anchored trained model ($t{+}1$ only, Audit 1):}} \\
Frozen correction head (zero) & 0.625 & -- & -- \\
Trained correction head (full) & 0.629 & -- & -- \\
\bottomrule
\end{tabular}
\end{table}

\begin{figure*}[t]
\centering
\includegraphics[width=0.82\textwidth]{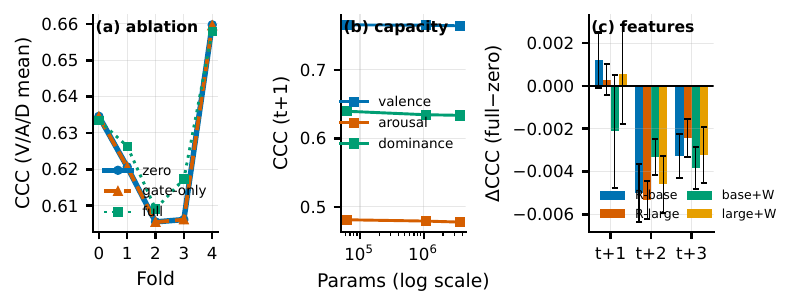}
\caption{Three audits. (a)~Freezing the correction head changes CCC by at most $0.02$ on all folds. (b)~A $60\times$ capacity sweep moves CCC in the third decimal (3 seeds). (c)~Across four feature configurations, the head's end-to-end CCC increment is negative with a bootstrap CI excluding zero in 23 of 24 cells at $t{+}2/t{+}3$.}
\label{fig:audit}
\end{figure*}

\phead{Audit 1 (ablation).} An architecture that \emph{anchors} its prediction on same-speaker persistence and learns only a gated correction matches its own frozen-head control ($t{+}1$: $0.6253$ frozen vs.\ $0.6288$ trained) to within $0.001$--$0.02$ CCC per fold: the correction head contributes nothing measurable beyond the persistence anchor it is built on.

\phead{Audit 2 (capacity).} Scaling the anchored model from $6.2{\times}10^{4}$ to $3.7{\times}10^{6}$ parameters moves 3-seed CCC by at most $0.0015$ per dimension.

\phead{Audit 3 (features).} Across RoBERTa-base/-large \cite{roberta}, each with and without WavLM \cite{wavlm} audio, the head's end-to-end CCC increment is noise at $t{+}1$ and significantly \emph{negative} at $t{+}2/t{+}3$, and conditional direction accuracy falls below chance ($0.42$--$0.44$, Fig.~\ref{fig:audit}).

\phead{Ruling out an annotation artifact -- partially.} IEMOCAP's sequential human VAD ratings \cite{iemocap} raise a confound: an annotator's self-anchoring, not real emotion dynamics, could drive the persistence floor. We test this directly by predicting one evaluator's rating from a \emph{different} evaluator's rating of the same speaker's prior utterance -- self-anchoring is physically impossible across annotators. Table~\ref{tab:annotator} reports this cross-annotator channel against a same-turn agreement ceiling. For \emph{valence}, cross-annotator persistence retains $85.1\%$ of the agreement ceiling (CCC $0.587$ against a ceiling of $0.689$ at gap 1, decaying to $73.9\%$ at gap 3): the signal is carried by what independent observers perceive, and self-anchoring is not a plausible explanation for the valence floor. For \emph{arousal} and \emph{dominance} the picture is materially weaker (ratio $39.9\%$ and $55.3\%$ at gap 1, falling to $13.1\%$ and $38.5\%$ at gap 3), so we do not claim the artifact hypothesis is ruled out for those dimensions -- only substantially weakened for valence, which also carries the largest share of the barrier's magnitude.

\begin{table}[t]
\caption{Annotator-artifact control (IEMOCAP, same-speaker gap 1, mean over V/A/D unless noted). Cross-annotator retains $68.4\%$ of the ceiling on average, concealing a per-dimension spread of $39.9$--$85.1\%$ at gap 1.}
\label{tab:annotator}
\centering
\begin{tabular}{lccc}
\toprule
Channel (gap$=1$) & CCC & $n$ pairs & Ratio to ceiling \\
\midrule
Mean-label persistence & 0.627 & 9737 & -- \\
Within-annotator & 0.620 & 20786 & -- \\
Cross-annotator & 0.281 & 24916 & 68.4\% \\
Agreement ceiling & 0.411 & 12456 & 100\% \\
\bottomrule
\end{tabular}
\end{table}

\phead{The Crack.} Departure \emph{occurrence} escapes the barrier: at threshold $q{=}75$, AUPRC exceeds the base rate ($0.237$) in fifteen of fifteen fold-model cells, up to $0.305$ with audio. Occurrence is learnable; magnitude and direction are not, under any tested configuration. The rest of the paper builds a forecasting system on the learnable side of this boundary.

\section{Commitment Policies}
\label{sec:policies}

\begin{table}[t]
\caption{Commitment policies, $\ltf$ (3-seed mean). a: calibrated threshold; b: cost-sensitive myopic; c: derailment-style delay rule; d: optimal stopping.}
\label{tab:policies}
\centering
\begin{tabular}{@{}llcccc@{}}
\toprule
Dataset & Model & a & b & c & d \\
\midrule
\multirow{2}{*}{DailyDialog} & GRU-MH & 0.337 & \textbf{0.343} & 0.336 & 0.304 \\
 & PsyDyn & 0.248 & \textbf{0.265} & 0.256 & 0.210 \\
\multirow{2}{*}{MELD} & GRU-MH & 0.542 & 0.552 & 0.543 & \textbf{0.635} \\
 & PsyDyn & 0.659 & 0.702 & 0.642 & \textbf{0.702} \\
\bottomrule
\end{tabular}
\end{table}

We compare four policies (Table~\ref{tab:policies}). Paired dialogue-level bootstrap (1{,}000 resamples) confirms the split: the optimal-stopping policy d beats the calibrated threshold on MELD by $+9.3$ points (GRU-MH) and $+4.3$ points (PsyDyn), yet \emph{loses} on DailyDialog by $3.3$--$3.8$ points, both with confidence intervals excluding zero. No policy dominates.

\begin{figure*}[t]
\centering
\includegraphics[width=0.85\textwidth]{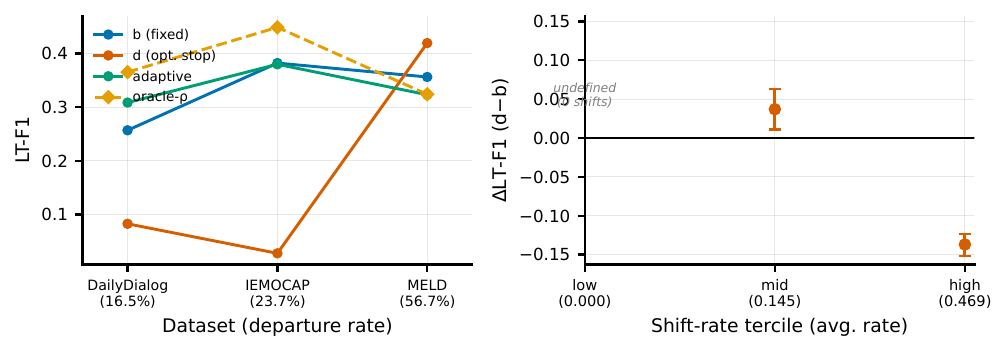}
\caption{Base rate is the switch, but not the whole story. Left: a rate-thresholded adaptive policy wins at low departure rates ($+5.1$ points over the fixed threshold on DailyDialog) and saturates at high rates, where optimal stopping takes over. Right: \emph{within} DailyDialog, optimal stopping loses most heavily in the highest-shift stratum ($-13.7$ points, CI excluding zero).}
\label{fig:policy}
\end{figure*}

\phead{Base rate is the switch.} A simpler adaptive policy that thresholds directly on an online estimate of the local departure rate $\rho_t$ recovers the better regime at each extreme: at low base rate (DailyDialog, $16.5\%$) it beats the fixed threshold by $+5.1$ points, with a further $+5.7$ points available to an oracle-$\rho$ upper bound; at high base rate (MELD, $56.7\%$) the oracle bound collapses to $+0.1$ points -- the local-rate feature saturates, and multi-step planning takes over. Within-dataset stratification complicates the story: inside DailyDialog, optimal stopping loses most heavily in the \emph{highest}-shift-rate stratum ($-13.7$ points), suggesting an interaction between base rate and confidence-trajectory quality.

\phead{A regime-adaptive meta-policy.} We train a small gate $g(\rho_t)\in[0,1]$ on an online estimate of the local departure rate, acting as the optimal-stopping policy with probability $g$ and the cost-sensitive policy otherwise. On DailyDialog the pooled meta-policy attains $\ltf=0.310$, ahead of the pooled best single policy ($0.304$) by $+0.6$ points; on MELD it matches the pooled best single policy within $0.06$ points. Zero-shot gate transfer across corpora is unreliable, so per-corpus calibration remains required; an oracle-$g$ bound retains $+1.7$--$+2.8$ points of headroom. CRC-wrapping the meta-policy keeps mean risk at or below target in all 12 cells on the main pipeline.

\begin{figure}[t]
\centering
\includegraphics[width=0.98\linewidth]{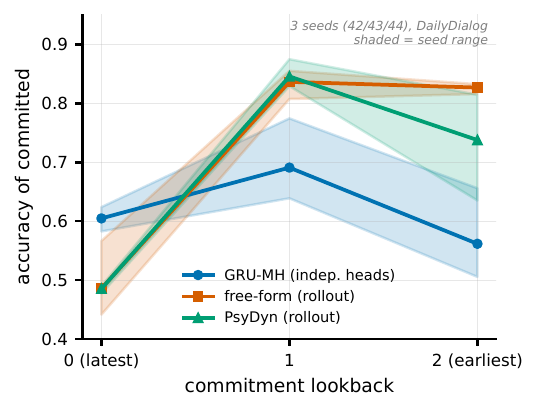}
\caption{Opposite earliness-accuracy profiles across architectures (3 seeds, mean line with seed-range shading, DailyDialog, single seed set).}
\label{fig:streaming}
\end{figure}

\phead{Mechanism evidence for policy-model co-design.} Increasing decision points from 3 to 5 systematically \emph{narrows} the gap between the optimal-stopping and myopic policies in both directions (DailyDialog: $-4.9\to-2.3$ points; MELD: $+11.2\to+6.9$ points), i.e., the planning advantage is largest when decisions are coarse. Architectures also exhibit \emph{divergent} earliness-accuracy profiles: shared-rollout heads self-select easy early commitments and gain with lead time in all 3 seeds, while independent multi-horizon heads gain far less and are sometimes more accurate at the last moment (Fig.~\ref{fig:streaming}) -- a policy tuned for one confidence-trajectory shape can be actively harmful for the other.

\section{Dialogue-Level Risk Control}
\label{sec:crc}

Given any fixed selective forecaster, we seek a distribution-free guarantee on the error rate of \emph{committed} predictions. Turns within a dialogue are strongly dependent, so we calibrate at the dialogue level. For dialogue $d$ with decision points $i=1,\dots,k_d$, base decisions $\mathrm{committed}_i$, confidences $c_i\in[\tfrac12,1]$, and correctness $\mathrm{corr}_i$, define a confidence veto $\mathrm{submit}_i(\lambda)=\mathrm{committed}_i\wedge(c_i\ge\lambda)$ and the ratio loss
\begin{equation}
L(d,\lambda)=\frac{\#\{i:\mathrm{submit}_i(\lambda),\neg\mathrm{corr}_i\}}
{\#\{i:\mathrm{submit}_i(\lambda)\}}\in[0,1],
\label{eq:loss}
\end{equation}
with $L=0$ if the dialogue fully abstains. Although submission sets are nested in $\lambda$, the \emph{ratio} loss is not monotone: raising $\lambda$ can first evict a correct low-confidence submission and raise the error ratio. We therefore work with the per-dialogue \emph{envelope} loss $\tilde L(d,\lambda)=\max_{\lambda'\ge\lambda}L(d,\lambda')$, which is non-increasing in $\lambda$ and dominates $L$ pointwise.

\begin{theorem}[Dialogue-level CRC]
\label{thm:crc}
Assume calibration dialogues $d_1,\dots,d_n$ and a test dialogue $d_{n+1}$ are exchangeable, and let $\tilde R(\lambda)=\tfrac1n\sum_{j\le n}\tilde L(d_j,\lambda)$ over a finite grid $\Lambda\subset[\tfrac12,1{+}\varepsilon]$ for some $\varepsilon>0$. Let $\hat\lambda=\inf\{\lambda\in\Lambda:\tfrac{n}{n+1}\tilde R(\lambda)+\tfrac{1}{n+1}\le\alpha\}$ (and $\hat\lambda=\max\Lambda$, forcing full abstention, if no such $\lambda$ exists). Then $\mathbb{E}[L(d_{n+1},\hat\lambda)]\le\mathbb{E}[\tilde L(d_{n+1},\hat\lambda)]\le\alpha$.
\end{theorem}

We state this as a Proposition, not a new theorem: it is a direct instantiation of the standard conformal-risk-control guarantee for monotone (or monotonized) losses \cite{crc} -- whose own non-monotone corollary already prescribes exactly the suffix-max monotonization $\tilde L$ we use -- applied to our dialogue-level exchangeable unit and ratio loss; the proof follows that argument and is included in Appendix A. What is specific to this paper is not the risk-control mechanism but its target: the dialogue-equal ratio loss, its abstention convention, its use inside a selective departure-forecasting pipeline, and the empirical risk--coverage--recall analysis of what the guarantee costs; the general non-monotone and selective-CRC theory is prior and concurrent work \cite{scrc,nonmonocrc}. The commitment policy is fit on validation dialogues only, and every CRC calibration/evaluation split is drawn exclusively from test dialogues that never participated in that fitting, as the Proposition's exchangeability assumption requires; MELD's dyadic test split has only 68 usable dialogues under this restriction, so its confidence intervals are correspondingly wide.

\begin{table*}[t]
\caption{Dialogue-level CRC, test-only calibration/evaluation splits (20 repeats per cell, per-dialogue-envelope calibrator). ``Frac.\ $\le\alpha$'' is the fraction of repeats individually at or below target; \emph{mean} risk is at or below $\alpha$ in all 12 cells. ``Cond.\ risk'' restricts to dialogues with at least one committed prediction (not what the Proposition bounds).}
\label{tab:crc}
\centering
\small
\begin{tabular}{@{}llccccccc@{}}
\toprule
Dataset & Model & $\alpha$ & Mean risk & 95\% CI & Cond.\ risk & Frac.\ $\le\alpha$ & Coverage \\
\midrule
\multirow{3}{*}{DailyDialog} & \multirow{3}{*}{GRU-MH} & 0.1 & 0.086 & [0.061,0.113] & 0.154 & 0.85 & 0.169 \\
 & & 0.2 & 0.166 & [0.127,0.209] & 0.217 & 0.95 & 0.308 \\
 & & 0.3 & 0.247 & [0.212,0.287] & 0.279 & 1.00 & 0.471 \\
\multirow{3}{*}{DailyDialog} & \multirow{3}{*}{PsyDyn} & 0.1 & 0.074 & [0.050,0.097] & 0.140 & 1.00 & 0.160 \\
 & & 0.2 & 0.156 & [0.133,0.185] & 0.179 & 1.00 & 0.388 \\
 & & 0.3 & 0.203 & [0.182,0.225] & 0.206 & 1.00 & 0.773 \\
\multirow{3}{*}{MELD} & \multirow{3}{*}{GRU-MH} & 0.1 & 0.077 & [0.000,0.184] & 0.570 & 0.75 & 0.021 \\
 & & 0.2 & 0.141 & [0.043,0.206] & 0.511 & 0.90 & 0.042 \\
 & & 0.3 & 0.144 & [0.073,0.206] & 0.516 & 1.00 & 0.043 \\
\multirow{3}{*}{MELD} & \multirow{3}{*}{PsyDyn} & 0.1 & 0.060 & [0.000,0.177] & 0.150 & 0.90 & 0.049 \\
 & & 0.2 & 0.165 & [0.059,0.294] & 0.248 & 0.50 & 0.085 \\
 & & 0.3 & 0.252 & [0.147,0.353] & 0.310 & 0.80 & 0.113 \\
\bottomrule
\end{tabular}
\end{table*}

\begin{figure*}[t]
\centering
\includegraphics[width=0.78\textwidth]{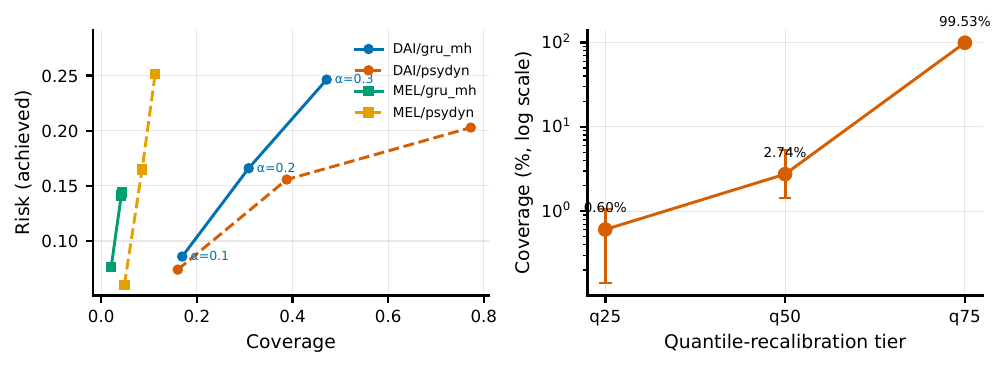}
\caption{Risk-coverage. Left: discrete tasks admit a tunable trade-off. Right: in the continuous setting the trade-off is nearly binary -- quantile-recalibrated $\alpha$ jumps coverage from under $5\%$ directly to $99.5\%$.}
\label{fig:crc}
\end{figure*}

\phead{Empirical verification and the price of the guarantee.} Across 2 datasets $\times$ 2 base models $\times$ $\alpha\in\{0.1,0.2,0.3\}$ (Table~\ref{tab:crc}), the \emph{mean} empirical risk across 20 test-only splits is at or below target in \textbf{12 of 12} cells. This is a claim about the mean over repeats, not about every individual repeat: on MELD's small test split, several individual repeats exceed $\alpha$ even where the mean does not. The guarantee comes at a measurable coverage cost -- $16$--$77\%$ on DailyDialog, only $2$--$11\%$ on MELD -- since the procedure partly controls risk through full abstention. Conditional risk (restricted to dialogues with at least one commitment) is substantially higher than the macro risk wherever abstention is heavy: on MELD, $0.51$--$0.57$ against a macro risk of only $0.08$--$0.14$, roughly a $7\times$ gap. ``Mean risk'' is what the Proposition's guarantee is about; ``conditional risk'' is what a user should expect the next time the system actually says something, and on MELD's small test split it is not small. In the continuous setting the trade-off is nearly binary (Fig.~\ref{fig:crc}, right): quantile-recalibrating $\alpha$ jumps coverage from under $5\%$ directly to $99.5\%$, with no tunable middle ground.

\phead{Risk on the alarms that matter.} Restricting to committed \emph{departure} alarms, a threshold recalibrated directly against the alarm-only loss achieves mean risk at or below $\alpha$ in \textbf{12 of 12} cells as well, so a practitioner who cares specifically about alarm risk gets the same style of guarantee by calibrating for it directly rather than reusing the pooled threshold. The cost is a severe recall penalty: on DailyDialog/GRU-MH, base recall $0.656$ becomes deployment recall $0.062$/$0.162$/$0.262$ at $\alpha{=}0.1$/$0.2$/$0.3$; on MELD/GRU-MH, base recall $0.530$ becomes $0.016$/$0.036$/$0.092$. A risk-controlled alarm system on these corpora is a high-precision, very-low-recall device: right about the few departures it commits to, but silent on most of them -- a trade-off, not risk alone, that a deployer must weigh.

\section{Experiments}
\label{sec:exp}

\subsection{Setup}

We evaluate on four corpora (Table~\ref{tab:datastats}): DailyDialog \cite{dailydialog} (13{,}118 dialogues; departure rate $17.3\%$), MELD \cite{meld} (1{,}432 dialogues; $55.4\%$), IEMOCAP \cite{iemocap} (151 dyadic dialogues, continuous VAD, 5-fold session-held-out), and EmoryNLP \cite{emorynlp} (897 scenes; $64.6\%$); Hi-EF \cite{hief} (53 dialogues) is reported separately in supplementary material available from the corresponding author upon request due to its small size. Because only PsyDyn's interpersonal-coupling mechanism currently supports more than two speakers, Tables~\ref{tab:policies}, \ref{tab:crc}, and~\ref{tab:auroc} evaluate all four model families on MELD's dyadic subset (643 of 1{,}432 dialogues, filtered to exactly two distinct speakers) using the same checkpoints and data loader; Table~\ref{tab:main} additionally reports PsyDyn on the full multi-party MELD via a separately trained mean-pooled coupling variant, marked with a dagger and not a same-data comparison against the other three families. Base models are GRU direct, GRU multi-horizon, free-form latent dynamics, and OU-coupled dynamics (PsyDyn), all built over frozen encoders (RoBERTa-base/-large \cite{roberta}; WavLM audio \cite{wavlm} where available); 3 seeds throughout. Splits are dialogue-level; leakage is asserted by a test suite. All experiments ran on one consumer laptop (CPU/MPS).

\begin{table}[t]
\caption{Dataset statistics. All splits are dialogue-level, verified by a leakage test suite.}
\label{tab:datastats}
\centering
\scriptsize
\setlength{\tabcolsep}{3pt}
\begin{tabular}{lccc}
\toprule
Dataset & Dialogues (tr/val/te) & Turns & Shift rate \\
\midrule
DailyDialog & 13118 (10494/1311/1313) & 102979 & 0.173 \\
MELD (mp) & 1432 (1145/143/144) & 13708 & 0.554 \\
Hi-EF & 53 (42/5/6) & 4779 & 0.640 \\
EmoryNLP & 897 (717/89/91) & 12606 & 0.646 \\
IEMOCAP (5-fold CV) & 151 & 10039 & N/A (VAD) \\
\bottomrule
\end{tabular}
\end{table}

\begin{table*}[t]
\caption{Departure forecasting across corpora (shift-F1). MELD is split into its dyadic subset (all four families) and the full multi-party corpus (PsyDyn only, marked $\dagger$). EmoryNLP is a 3-seed mean$\pm$std; bold marks the best departure forecaster where its shift-F1 exceeds Random@prior on the same pool.}
\label{tab:main}
\centering
\small
\begin{tabular}{@{}lcccc@{}}
\toprule
Model & DailyDialog & MELD (dyadic) & MELD (multi-party)$^{\dagger}$ & EmoryNLP (3-seed) \\
\midrule
Random@prior & 0.179 & 0.581 & 0.581$^{\dagger}$ & $0.609\pm0.025$ \\
GRU direct & \textbf{0.393} & 0.685 & --$^{\dagger}$ & $\mathbf{0.737\pm0.031}$ \\
GRU multi-horizon & 0.387 & 0.697 & --$^{\dagger}$ & $0.673\pm0.088$ \\
Free-form dynamics & 0.389 & \textbf{0.710} & --$^{\dagger}$ & $0.673\pm0.150$ \\
OU-coupled (PsyDyn) & 0.371 & 0.688 & 0.665$^{\dagger}$ & $0.665\pm0.117$ \\
\bottomrule
\end{tabular}
\end{table*}

\subsection{Occurrence Learnability Tracks the Base Rate Across Corpora}

Shift-F1 against a majority-class base rate is misleading on EmoryNLP: an always-positive rule scores shift-F1 $0.76$ with zero ranking ability, a base-rate artifact. We instead read occurrence learnability with threshold-free AUROC (Table~\ref{tab:auroc}), which shows a monotone association across the three corpora we test: AUROC drops from $0.72$ to $0.56$ to $0.50$ as the base rate climbs from $0.17$ to $0.55$ to $0.65$ (DailyDialog / MELD / EmoryNLP). We describe this as a cross-corpus empirical association, not a law: three data points cannot establish a causal or statistical law, and the three corpora differ simultaneously in label taxonomy, source domain, dialogue length, speaker count, and annotation protocol. Within this scope, occurrence is \emph{most} learnable in the low-base-rate regime where anticipation matters most, and the one chance-level corpus (EmoryNLP) is consistent with, rather than a counterexample to, the trend.

\begin{table}[t]
\caption{Threshold-free occurrence learnability (3 seeds each). AUROC/MCC are 0.5/0 for both baselines by construction; EmoryNLP is indistinguishable from chance across the full model family.}
\label{tab:auroc}
\centering
\small
\begin{tabular}{llcc}
\toprule
Dataset & Model & AUROC [range] & MCC \\
\midrule
DailyDialog & GRU-direct & 0.720 [0.717,0.726] & 0.247 \\
DailyDialog & GRU-MH & 0.716 [0.713,0.721] & 0.239 \\
DailyDialog & Free-form & 0.720 [0.719,0.722] & 0.242 \\
DailyDialog & PsyDyn & 0.687 [0.683,0.689] & 0.220 \\
MELD & GRU-direct & 0.529 [0.497,0.556] & 0.031 \\
MELD & GRU-MH & 0.516 [0.492,0.545] & 0.057 \\
MELD & Free-form & 0.541 [0.507,0.561] & 0.033 \\
MELD & PsyDyn & 0.563 [0.549,0.585] & 0.091 \\
EmoryNLP & GRU-direct & 0.495 [0.488,0.506] & 0.028 \\
EmoryNLP & GRU-MH & 0.504 [0.485,0.542] & $-0.013$ \\
EmoryNLP & Free-form & 0.497 [0.490,0.501] & $-0.023$ \\
EmoryNLP & PsyDyn & 0.463 [0.456,0.475] & $-0.011$ \\
\bottomrule
\end{tabular}
\end{table}

\subsection{Architecture Dependence, Abstention, and LLM Confidence}

A shallow, non-recurrent baseline (logistic regression / small MLP over a single frozen embedding, no dialogue history, no recurrence) trained on the true same-speaker anchor remains below every trained architecture on DailyDialog (AUROC $0.672$/$0.643$ versus $0.687$--$0.720$), but \emph{exceeds} every trained architecture on MELD (logistic regression AUROC $0.587$, MCC $0.150$, versus $0.516$--$0.563$/$0.031$--$0.091$). The trained architectures' advantage over a simple classifier is itself base-rate- and corpus-dependent -- real but modest on DailyDialog, and on MELD, where the departure rate is high and the occurrence signal is weak for everyone, the added complexity does not clearly pay for itself. At 70\% coverage, calibrated abstention improves shift-F1 in \textbf{27 of 28} model-dataset-capacity cells across a base/512/1024 parameter sweep on DailyDialog/MELD ($+1.35$ to $+5.38$ points) and 3 of 4 cells on EmoryNLP; the lone exception, free-form dynamics on EmoryNLP ($-6.1$ points), is exactly the chance-AUROC corpus. Full extended tables for this sweep are available from the corresponding author upon request.

\begin{figure}[t]
\centering
\includegraphics[width=0.98\linewidth]{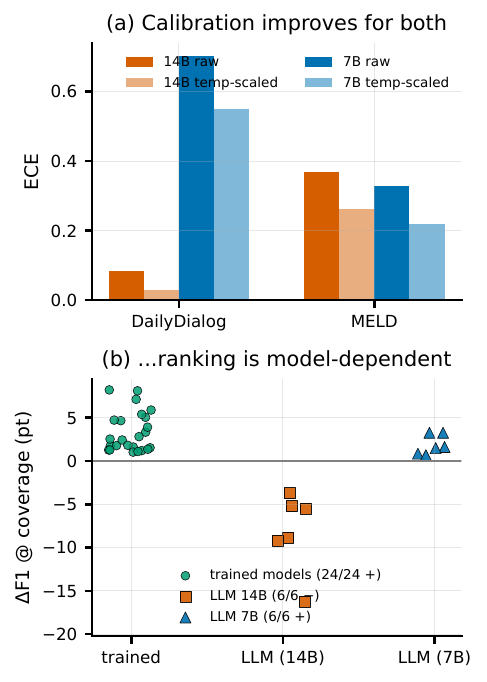}
\caption{Calibrated confidence does not imply reliable ranking, for at least one of the two LLMs tested. Top: temperature scaling improves ECE for both. Bottom: abstention by the 14B model's confidence hurts in 6/6 cells; by the 7B model's, it helps in 6/6.}
\label{fig:llm}
\end{figure}

Two instruction-tuned open-weight LLMs from different parameter-count families (\texttt{qwen3:14b} and \texttt{mistral:7b}, served locally, no external API calls) are prompted zero-shot to predict whether the named target speaker's next turn will depart from their own previous state, together with a self-reported confidence; the 14B model was queried on the full DailyDialog/MELD test sets and the 7B model on a fixed 500-target subsample (Table~\ref{tab:llm-full}, Fig.~\ref{fig:llm}). Temperature scaling improves expected calibration error (ECE) for both models by $65\%/29\%$ (14B) and $22\%/33\%$ (7B), but the two diverge sharply on \emph{ranking}: abstention by the 14B model's confidence \emph{hurts} in \textbf{6 of 6} cells ($-3.7$ to $-16.2$ points); by the 7B model's, it \emph{helps} in \textbf{6 of 6} ($+0.7$ to $+3.3$ points). Because the two models differ simultaneously in parameter count, model family, and evaluation-set size, scale and family are confounded, and we state the finding as \emph{model}-dependent, not family-dependent: at least one widely used LLM fails this ranking test, while at least one other passes it. A replication under error-framing and a two-step decoupled confidence protocol (supplementary material available from the corresponding author upon request) reproduces the same split between the two models, ruling out a single-protocol artifact as the explanation. Trained forecasters, by contrast, are positive in every cell we test.

\begin{table}[t]
\caption{Both LLMs tested: full metrics. The 14B model's ranking deltas are negative in 6 of 6 cells; the 7B model's are positive in 6 of 6 cells.}
\label{tab:llm-full}
\centering
\scriptsize
\setlength{\tabcolsep}{3.2pt}
\begin{tabular}{llcccccc}
\toprule
Family & Dataset & Shift-F1 & ECE raw & ECE cal. & $\Delta$50\% & $\Delta$70\% & $\Delta$90\% \\
\midrule
14B & DailyDialog & 0.101 & 0.084 & 0.029 & $-8.84$ & $-9.22$ & $-5.48$ \\
14B & MELD & 0.334 & 0.368 & 0.261 & $-16.24$ & $-5.22$ & $-3.72$ \\
7B & DailyDialog & 0.274 & 0.704 & 0.549 & $+3.26$ & $+3.26$ & $+0.85$ \\
7B & MELD & 0.719 & 0.330 & 0.221 & $+1.62$ & $+1.49$ & $+0.67$ \\
\bottomrule
\end{tabular}
\end{table}

Finally, the specific discretized OU-plus-coupling parameterization we implement underperforms an encoder-matched GRU baseline both discretely ($-4.7$ points shift-F1) and continuously (worse change-segment CCC in 15 of 15 fold-horizon cells). We do not read this as falsifying the psychological theory it is inspired by \cite{oravecz2011,butler2011} -- only this specific implementation as a \emph{value} forecaster; its internal ablation ordering is stable and consistent with the theory's own emphasis, since removing interpersonal coupling hurts far more than sharing the inertia parameter across speakers.

\section{Design Principles}
\label{sec:principles}

\textbf{P1: When a reliable historical emotion-state observation is available, report persistence and state-space baselines and measure incremental value beyond them.} Same-speaker persistence is the floor and a classical causal Kalman filter over the same observed anchor is the closest thing we find to a ceiling; every unconstrained neural model we train falls well below both. We do \emph{not} generalize this to ``don't fight persistence'' as blanket advice, since the ordering depends on state observability: when the anchor must be estimated rather than observed, estimated persistence drops below even a direct neural forecaster. The robust recommendation is evaluative, not architectural: whenever a trustworthy historical state is available, report persistence and a state-space baseline over it, and evaluate learned models on the value they add \emph{beyond} those references.

\textbf{P2: Within the tested state-observed setting, prioritize forecasting departure occurrence over magnitude.} Occurrence is learnable while direction and size are not, under ablation, capacity, and multimodal sweeps, and occurrence learnability tracks the departure base rate across the three corpora we test -- an empirical association, not an established law, and near chance on EmoryNLP -- scaling inversely with it, so anticipation appears strongest in the rare-event regime that makes forecasting worth doing.

\textbf{P3: Never proxy ranking by calibration.} A model can be well-calibrated and still order its own errors worse than chance; selective deployment must test ranking directly via risk-coverage curves and, where stakes demand it, wrap commitment in distribution-free risk control rather than trusting verbalized or softmax confidence alone.

\section{Discussion}
\label{sec:discussion}

\subsection{Limitations}

Our strongest negative results are bounded by the configurations tested: frozen pretrained encoders with trained heads up to $3.7$M parameters, evaluated on four corpora. They are also bounded by the oracle-anchor assumption of Section~\ref{sec:barrier}: same-speaker persistence's dominance collapses once its anchor must be estimated rather than observed, so the Persistence Barrier is a statement about \emph{state-observed} forecasting, not a fully realistic pipeline. We do not test full end-to-end fine-tuning, larger Transformer-based sequence models, or a directly-CCC-optimized regression head, and cannot rule out that one of these would close some of the gap. Several claims in this paper are stated as counts over many cells (e.g., ``15 of 15 fold-model cells''); these are not independent replications, since multiple seeds and folds share the same underlying dialogues and encoder, and should be read as describing consistency across our experimental grid rather than additional independent statistical evidence -- confidence intervals, where reported, use dialogue- or session-level bootstrap rather than treating seeds as i.i.d.\ samples. All dialogue-level splits are re-derived at an approximately 80/10/10 ratio with a fixed seed rather than the original DailyDialog/MELD releases' own splits, so our numbers are not directly comparable to prior work on the original splits; an episode-disjoint re-split of MELD and EmoryNLP (supplementary material available from the corresponding author upon request) moves shift-F1 by at most a few points per cell, though recurring-character-level leakage across episodes is not tested and remains an open question. Our architectures assume the target position's speaker identity is known at commit time, a real but modest assumption on IEMOCAP and MELD's dyadic subset and the most consequential on the full multi-party corpora, where the next speaker is a genuine scheduling fact we do not attempt to forecast jointly with the emotional departure. The mean-then-envelope CRC calibrator, reported alongside the Proposition-covered envelope-then-mean calibrator as a secondary comparison, fails to control mean risk in 1 of 12 main-pipeline cells in supplementary material available from the corresponding author upon request, which is itself part of why we do not treat it as the primary claim.

\subsection{Ethical and Societal Considerations}

Anticipating a person's emotional state before it is expressed raises surveillance concerns under emerging regulation. We scope this precisely: the EU AI Act (Art.~3(39)) defines an ``emotion recognition system'' as one inferring emotion from \emph{biometric} data, and Art.~5(1)(f) restricts such inference specifically in workplace/education contexts -- it does not categorically cover text-only inference, which is most of our experiments. This is precisely why we frame the deployable output of this work as \emph{selective}: our risk-controlled abstention procedure makes ``decline to predict'' a first-class, statistically licensed output, so that a deployed system comes with an auditable, finite-sample bound on dialogue-averaged macro loss (not a bound on the conditional error among submitted predictions) instead of an unbounded average error rate. We recommend that any deployment in emotionally sensitive settings be accompanied by disclosure to the monitored party, human-in-the-loop review of high-stakes committed predictions, and periodic re-calibration of the CRC threshold against a held-out population.

\subsection{Practical Deployment Implications}

Three findings bear directly on system design. First, because occurrence learnability degrades as the departure base rate rises, a deployed system should measure its target population's local departure rate before committing to a forecasting-based intervention, rather than assuming the benefit transfers. Second, because the best commitment policy is base-rate-dependent and zero-shot policy transfer across domains is unreliable, a policy trained on one deployment's data should not be assumed to transfer without re-calibration, even within the same architecture family. Third, because LLM verbalized confidence is an unreliable ranking signal for at least one widely used model, and reliability depends on the confidence-elicitation protocol, any system that relies on an LLM's self-reported confidence to gate action should validate that signal empirically before deployment.

\section{Conclusion}
\label{sec:conclusion}

Within the frozen-encoder, state-observed testbed we audit, residual value forecasting beyond an observed emotional state is barricaded by same-speaker persistence -- a barrier that itself dissolves once the state must be estimated rather than observed, which limits this conclusion to the state-observed setting. On the other side of it, the task we find tractable and useful is selective, risk-controlled anticipation of departures. We provide the formalization, the metric, the policies, the guarantee, and the audited evidence, including the negative results and their disclosed scope boundaries, on which future affective forecasting systems can be built and tested against stronger baselines. All code, configurations, and result logs needed to recompute every number and regenerate every figure are available from the corresponding author upon request; the proof of Proposition~\ref{thm:crc} is included in Appendix A, and implementation details and extended per-cell tables are available from the corresponding author upon request.

\section*{Acknowledgment}

The authors used a large language model assistant to help with English-language editing, LaTeX formatting, code review, and the documentation and presentation of supplementary analyses. All experimental designs, executions, interpretations, and conclusions were independently reviewed and verified by the authors, who assume full responsibility for the submitted work.

\section*{Appendix A: Proof of Dialogue-Level Risk Control}
Fix the finite grid $\Lambda=\{\lambda_1<\dots<\lambda_m\}$ and recall
$\tilde L(d,\lambda)=\max_{\lambda'\ge\lambda}L(d,\lambda')$, so that for
every dialogue $d$: (i) $\tilde L(d,\cdot)$ is non-increasing on
$\Lambda$; (ii) $0\le L(d,\lambda)\le\tilde L(d,\lambda)\le 1$; and
(iii) at $\lambda=\max\Lambda>1$ every dialogue fully abstains, so
$L(d,\max\Lambda)=\tilde L(d,\max\Lambda)=0$.

Consider the augmented (exchangeable) collection $d_1,\dots,d_{n+1}$ and
define the augmented empirical risk
\[
\tilde R_{n+1}(\lambda)\;=\;\frac{1}{n+1}\sum_{j=1}^{n+1}
\tilde L(d_j,\lambda),
\]
non-increasing in $\lambda$ as an average of non-increasing functions.
Define the augmented threshold
$\hat\lambda_{n+1}=\inf\{\lambda\in\Lambda:\tilde
R_{n+1}(\lambda)\le\alpha\}$; by property (iii),
$\tilde R_{n+1}(\max\Lambda)=0\le\alpha$, so the feasible set is
nonempty and the infimum is attained on the finite grid.

\begin{lemma}
\label{lem:order}
$\hat\lambda\ \ge\ \hat\lambda_{n+1}$ almost surely, where $\hat\lambda$
is the calibration-only threshold of Proposition~\ref{thm:crc}.
\end{lemma}

\begin{proof}
For any $\lambda\in\Lambda$,
\[
\tilde R_{n+1}(\lambda)
=\frac{n}{n+1}\tilde R(\lambda)+\frac{1}{n+1}\tilde
L(d_{n+1},\lambda)
\;\le\;\frac{n}{n+1}\tilde R(\lambda)+\frac{1}{n+1},
\]
using $\tilde L\le 1$. Hence whenever the calibration condition
$\frac{n}{n+1}\tilde R(\lambda)+\frac{1}{n+1}\le\alpha$ holds,
$\tilde R_{n+1}(\lambda)\le\alpha$ also holds: the feasible set defining
$\hat\lambda$ is contained in that defining $\hat\lambda_{n+1}$. Both
infima exist on the finite grid (the latter's feasible set is nonempty by
property (iii); if the former's is empty, $\hat\lambda=\max\Lambda\ge
\hat\lambda_{n+1}$ trivially). Infima over nested sets are ordered:
$\hat\lambda\ge\hat\lambda_{n+1}$.
\end{proof}

\begin{proof}[Proof of Proposition~\ref{thm:crc}]
By monotonicity of $\tilde L(d_{n+1},\cdot)$ and Lemma~\ref{lem:order},
$\tilde L(d_{n+1},\hat\lambda)\le\tilde L(d_{n+1},\hat\lambda_{n+1})$,
so it suffices to bound
$\mathbb{E}[\tilde L(d_{n+1},\hat\lambda_{n+1})]$.

The threshold $\hat\lambda_{n+1}$ is a symmetric function of the multiset
$\{d_1,\dots,d_{n+1}\}$: it depends on the dialogues only through the
augmented risk curve $\tilde R_{n+1}(\cdot)$, which is permutation
invariant. By exchangeability, conditionally on the multiset the identity
of the test dialogue is uniform over the $n+1$ elements; hence
\[
\mathbb{E}\!\left[\tilde L(d_{n+1},\hat\lambda_{n+1})\right]
=\mathbb{E}\!\left[\frac{1}{n+1}\sum_{j=1}^{n+1}
\tilde L(d_j,\hat\lambda_{n+1})\right]
=\mathbb{E}\!\left[\tilde R_{n+1}(\hat\lambda_{n+1})\right]
\le\alpha,
\]
where the final inequality holds deterministically because
$\hat\lambda_{n+1}$ is, by construction, a feasible point of its defining
condition on the finite grid. Combining,
$\mathbb{E}[\tilde L(d_{n+1},\hat\lambda)]\le\alpha$, and finally
$\mathbb{E}[L(d_{n+1},\hat\lambda)]\le
\mathbb{E}[\tilde L(d_{n+1},\hat\lambda)]\le\alpha$ by pointwise
dominance. \qedhere
\end{proof}

\end{document}